\documentclass[a4paper, 10pt, conference]{ieeeconf}      
\usepackage{FG2018}
\usepackage{times}
\usepackage{epsfig}
\usepackage{graphicx}
\usepackage{amsmath}
\usepackage{amssymb}
\usepackage{color}
\usepackage{url}
\usepackage{float}
\usepackage{caption}
\usepackage{subcaption}

\FGfinalcopy 

\IEEEoverridecommandlockouts                              
\def\FGPaperID{185} 

\title{\LARGE \bf
End-to-end Learning for 3D Facial Animation\\ from Raw Waveforms of Speech
}

\author{\parbox{16cm}{\centering
    {\large Hai X. Pham, Yuting Wang, Vladimir Pavlovic}\\
    {\large \textit{Department of Computer Science, Rutgers University}}\\
	{\large \textit{\{hxp1,yw632,vladimir\}@cs.rutgers.edu}}}
}

\begin{document}

\ifFGfinal
\thispagestyle{empty}
\pagestyle{empty}
\else
\author{Anonymous FG 2018 submission\\ Paper ID \FGPaperID \\}
\pagestyle{plain}
\fi
\maketitle

%


\begin{abstract}

We present a deep learning framework for real-time speech-driven 3D facial animation
from just raw waveforms. Our deep neural network directly maps an input sequence of 
speech audio to a series of micro facial action unit activations and head rotations 
to drive a 3D blendshape face model. In particular, our deep model is able to learn
the latent representations of time-varying contextual information and affective
states within the speech. Hence, our model not only activates appropriate 
facial action units at inference to depict different utterance generating 
actions, in the form of lip movements, but also, without any assumption, 
automatically estimates emotional intensity of the speaker and reproduces 
her ever-changing affective states by adjusting strength of facial unit activations.
For example, in a happy speech, the mouth opens wider than normal, while other
facial units are relaxed; or in a surprised state, both eyebrows raise higher.
Experiments on a diverse audiovisual corpus of different actors across a wide range
of emotional states show interesting and promising results of our approach. 
Being speaker-independent, our generalized model is readily applicable to various tasks
in human-machine interaction and animation.

\end{abstract}

\section{INTRODUCTION}
\label{sec:intro}

Face synthesis is essential to many applications, such as computer games, 
animated movies, teleconferencing, talking agents, etc. Traditional facial capture 
approaches have gained tremendous successes, reconstructing high 
level of realism.  
Yet, active face capture rigs utilizing motion sensors/markers are
expensive and time-consuming to use. Alternatively, passive techniques capturing facial
transformations from cameras, although less accurate,
have achieved very impressive performance.

There lies one problem with vision-based facial capture approaches, however,
where part of the face is occluded, e.g. when a person is wearing a mixed reality visor,
or in the extreme situation where the entire visual appearance is non-existent. 
In such cases, other input modalities, such as audio, may be exploited to infer facial actions.
Indeed, research on speech-driven face synthesis has regained attention of the community
in recent time. Latest works
~\cite{karras_sigg2017,pham_cvprw2017,supasorn_sigg2017,taylor_sigg2017}
employ deep neural networks in order to model the highly non-linear mapping from 
input speech domain, either as audio or phonemes, to visual facial features. Particularly,
in approaches by Karras et al.~\cite{karras_sigg2017} and Pham et al.~\cite{pham_cvprw2017},
the reconstruction of facial emotion is also taken into account to generate fully transform 
3D facial shapes. The method in~\cite{karras_sigg2017} explicitly specifies the emotional 
state as an additional input beside waveforms, whereas~\cite{pham_cvprw2017} implicitly infers
affective states from acoustic features, and represents emotions via blendshape weights~\cite{cao_fwh2013gy}.

In this work, we further improve the approach of~\cite{pham_cvprw2017} in several ways, in order
to recreate a better 3D talking avatar that can naturally rotate and perform micro facial actions
to represent the time-varying contextual information and emotional intensity from speech in real-time.
Firstly, we forgo using handcrafted, high-level acoustic features such as chromagram or mel-frequency
cepstral coefficients (MFCC), which, as the authors conjectured, may cause the loss of important 
information to identify some specific emotions, e.g. happy. 
Instead, we directly use Fourier transformed spectrogram as input to our neural network. 
Secondly, we employ convolutional neural networks (CNN) to learn meaningful acoustic feature 
representations, and take advantage of the locality and shift invariance in the frequency 
domain of audio signal.
Thirdly, we combine these convolutional layers with recurrent layer in an end-to-end
framework, which learns both temporal transition of facial movements, as well as spontaneous
actions and varying emotional states from only speech sequences. Experiments on the RAVDESS
audiovisual corpus~\cite{ravdess} demonstrate promising results of our approach in real-time speech-driven
3D facial animation.

The organization of the paper is as follows. Section~\ref{sec:relwk} summarizes other studies related
to our work. Our approach is explained in details in Section~\ref{sec:frmwk}. Experiments are described
in Section~\ref{sec:eval}, before Section~\ref{sec:conclude} concludes our work.

\section{Related Work}
\label{sec:relwk}
\textit{"Talking head"}, is a research topic where an avatar is animated to imitate human talking.
Various approaches have been developed to synthesize a face model driven by either 
speech audio~\cite{fan_mta16,xie_multimed07,sako_icslp00} or
transcripts~\cite{wang_sandbox07,cosatto_03}. 
Essentially, every talking head animation technique develops a mapping from an input speech
to visual features, and can be formulated as a classification or regression task.
Classification approaches usually identify phonetic unit (phonemes) from speech and map to visual
units (visemes) based on specific rules, and animation is generated by morphing these key images.
On the other hand, regression approaches can directly generate visual parameters and their trajectories
from input features. Early research on talking head used Hidden Markov Models (HMMs) with some successes
~\cite{wang_interspk2010,wang_interspk11}, despite certain limitations of HMM framework such as
oversmoothing trajectory. 

In recent years, deep neural networks have been successfully applied to speech synthesis
~\cite{qian_icassp14,zen_icassp13} and facial animation~\cite{ding_mta15,zhang_interspk13,fan_mta16}
with superior performance. This is because deep neural networks (DNN) are able to learn the correlation 
of high-dimensional input data, and, in case of recurrent neural network (RNN), 
long-term relation, as well as the highly non-linear mapping between input and output features. 
Taylor et al.~\cite{taylor_sigg2017} propose a system using DNN to estimate active appearance model (AAM) coefficients 
from input phonemes, which can be generalized well to different speeches and languages, and 
face shapes can be retargeted to drive 3D face models. 
Suwajanakorn et al.~\cite{supasorn_sigg2017} use long short-term memory (LSTM) 
RNN to predict 2D lip landmarks from input acoustic features, which are used to
synthesize lip movements. 
Fan et al.~\cite{fan_mta16} use both acoustic and text features to estimate active appearance model AAM coefficients
of the mouth area, which then be grafted onto an actual image to produce a photo-realistic talking head.
Karras et al.~\cite{karras_sigg2017} propose a deep convolutional neural network (CNN) that jointly
takes audio autocorrelation coefficients and emotional state to output an entire 3D face shape.

In terms of the underlying face model, these approaches can be categorized into
image-based~\cite{bregler_siggraph07,cosatto_03,ezzat_siggraph02,wang_interspk2010,xie_multimed07,fan_mta16}
and model-based~\cite{blanz_99,blanz_03,salvi_eurasip09,wu_interspk06,ding_mta15,yongcao_tog05} approaches.
Image-based approaches compose photo-realistic output by concatenating short clips, 
or stitch different regions from a sample database together. 
However, their performance and quality are limited by the amount of samples in the database, 
thus it is difficult to generalize to a large corpus of speeches, which would require 
a tremendous amount of image samples to cover all possible facial appearances.
In contrast, although lacking in photo-realism, model-based approaches enjoy the flexibility
of a deformable model, which is controlled by only a set of parameters, and more straightforward
modeling. Pham et al.~\cite{pham_cvprw2017} propose a mapping from acoustic features 
to blending weights of a blendshape model~\cite{cao_fwh2013gy}. This face model allows
emotional representation that can be inferred from speech, without explicitly defining the emotion
as input, or artificially adding emotion to the face model in postprocessing. Our approach also enjoys
the flexibility of blendshape model in 3D face reconstruction from speech.

\textit{CNN-based speech modeling}. Convolutional neural networks~\cite{lecun_cnn} have achieved great successes
in many vision tasks e.g. image classification or segmentation. Their efficient filter design
allows deeper network, enables learning features from data directly 
while being robust to noise and small shift,
thus usually having better performance than prior modeling techniques. In recent years, CNNs have been also
employed in speech recognition tasks, that directly model the raw waveforms by
taking advantage of the locality and translation invariance in 
time~\cite{trigeorgis_is2016,palaz_is2013,hoshen_icassp2015} and frequency domain
~\cite{deng_icassp2013,hamid_icassp2012,hamid_TASLP2014,sainath_NN2015,sainath_icassp2015,sainath_is2015}.
In this work, we also employ convolutions in the time-frequency domain,
and formulate an end-to-end deep neural network that directly maps input waveforms 
to blendshape weights.

\section{DEEP END-TO-END LEARNING\\ FOR 3D FACE SYNTHESIS FROM SPEECH}
\label{sec:frmwk}
\subsection{Face Representation}
\label{sec:face}
Our work makes use of the 3D blendshape face model from the FaceWarehouse database~\cite{cao_fwh2013gy}, which has been utilized successfully in visual 3D face tracking tasks~\cite{pham_icpr2016,pham_3dv2016}.
An arbitrary fully transformed facial shape $S$ can be composed as:
\begin{equation}
S = R\left( {{B_0} + \sum\limits_{i = 1}^N {({B_i} - {B_0}){e_i}} } \right),
\label{eq:full_shape}
\end{equation}
where $(R,e)$ are rotation and expression blending parameters, respectively, 
$\{B_i | i=1..N\}$ are personalized expression blendshape bases of a particular person,
and $\{e_i\}$ are constrained within $[0,1]$. $B_0$ is the neutral posed blendshape.

Similar to~\cite{pham_cvprw2017}, our deep model also generates $(R,e)$, where $R$ is
represented by three free parameters of a quaternion, and $e$ is a vector of length $N=46$.
We use the 3D face tracker in~\cite{pham_icpr2016} to extract these parameters from training
videos.

\begin{figure}[!ht]
	\centering
	\includegraphics[width=8cm]{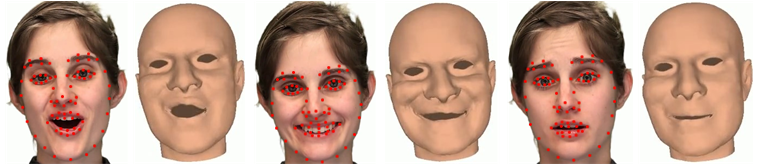}
	\caption{A few samples from the RAVDESS database, where a 3D facial blendshape (right)
		is aligned to the face of the actor (left) in the corresponding frame. 
		Red dots indicate 3D landmarks of the model projected to the image plane.}
	\label{fig:face}
\end{figure}

\subsection{Framework Architecture}
\label{sec:overview}

\begin{figure*}[!ht]
	\centering
	\includegraphics[width=\linewidth]{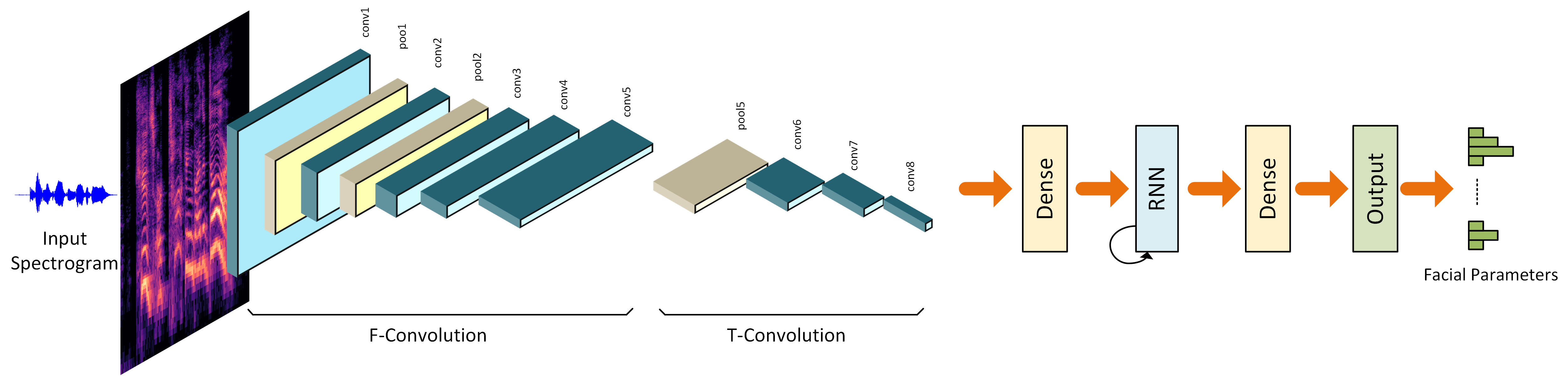}
	\caption{The proposed end-to-end speech-driven 3D facial animation framework.
		The input spectrogram is first convolved over frequency axis (F-convolution), 
		then over time (T-convolution). Detailed network architecture is 
		described in Table~\ref{tbl:config}.}
	\label{fig:fw}
\end{figure*}

Our end-to-end deep neural network is illustrated in Figure~\ref{fig:fw}. 
The input to our model is raw time-frequency spectrograms of audio signals.
Specifically, each spectrogram contains 128 frequency power bands across
32 time frames, in the form of a 2D (frequency-time) array suitable for
CNN. We apply convolutions on frequency and time separately, similar to~\cite{karras_sigg2017, sainath_is2015}, as this practice has been empirically
shown to reduce overfitting, furthermore, using smaller filters requires less computation, 
which consequently speeds up the training and inference. The network architecture is detailed
in Table~\ref{tbl:config}. Specifically, the input spectrogram is first convolved and pooled on the
frequency axis with the downsampling factor of two, until the frequency dimension is reduced to one.
Then, convolution and pooling is applied on the time axis. A dense layer is placed on top
of CNN, which feeds to a unidirectional recurrent layer. In this work, we report the model performance
where the recurrent layer utilizes either LSTM~\cite{lstm_97} 
or gated recurrent unit (GRU)~\cite{gru_chung_nipsw14} cells.
The output from RNN is passed to another dense layer,
whose purpose is to reduce the oversmoothing tendency of RNN and to allow more spontaneous changes
in facial unit activations. Every convolutional layer is followed by a batch normalization layer,
except the last one (\textit{Conv8} in Table~\ref{tbl:config}).

\begin{table}[!ht]
	\centering
	\caption{Configuration of our hidden neural layers.}
	\label{tbl:config}
	\begin{tabular}{|l|c|c|c|c|}
		\hline
		Name	& Filter Size & Stride & Hidden Layer Size & Activation	\\\hline
		Input	&		&		&	$128 \times 32$	&	 \\\hline\hline
		Conv1	&	(3,1) & (2,1) & $64 \times 64 \times 32$ & ReLU \\\hline
		Pool1	&	(2,1) & (2,1) & $64 \times 32 \times 32$ & \\\hline
		Conv2	&	(3,1) & (2,1) & $96 \times 16 \times 32$ & ReLU \\\hline
		Pool2	&	(2,1) & (2,1) & $96 \times 8 \times 32$ & \\\hline
		Conv3	&	(3,1) & (2,1) & $128 \times 4 \times 32$ & ReLU \\\hline
		Conv4	&	(3,1) & (2,1) & $160 \times 2 \times 32$ & ReLU \\\hline
		Conv5	&	(2,1) & (2,1) & $256 \times 1 \times 32$ & ReLU \\\hline\hline
		Pool5	&	(1,2) & (1,2) & $256 \times 1 \times 16$ & \\\hline
		Conv6	&	(1,3) & (1,2) & $256 \times 1 \times 8$ & ReLU \\\hline
		Conv7	&	(1,3) & (1,2) & $256 \times 1 \times 4$ & ReLU \\\hline
		Conv8	&	(1,4) & (1,4) & $256 \times 1 \times 1$ & ReLU \\\hline\hline
		Dense1	&		&		&	256 & tanh \\\hline
		RNN		&		&		&	256 & \\\hline
		Dense2	&		&		&	256 & tanh \\\hline
		Output 	&		&		&	49 & \\\hline
	\end{tabular}
\end{table}

\subsection{Training Details}

\textit{Audio processing}. For each video frame $t$ in the corpus, we extract a 96ms audio
frame sampled at 44.1kHz,
including the acoustic data of the current frame and the previous frames. With the intended
application in real-time animation, we do not consider any delay to gather future data, 
i.e. audio samples of frames $t+1$ onward, as they are unknown in a live streaming scenario.
Instead, temporal transition will be modeled by the recurrent layer.
We apply FFT with window size of 256 and hop length of 128, to recover
a power spectrogram of 128 frequency bands and 32 time frames.

\textit{Loss function}. Our framework maps input sequence of spectrograms $x_t,t=1..T$
to output sequence of shape parameter vectors $y_t$, where $T$ is the number of video frames. 
Thus, at any given time $t$, the deep model estimates $y_t = ({R_t}, {e_t})$ 
from an input spectrogram $x_t$. Similar to~\cite{pham_cvprw2017},
we split the output into two separate layers, $Y_R$ for rotation and $Y_e$ for expression weights.
$Y_R$ has \textit{tanh} activation to c, whereas $Y_e$ uses \textit{sigmoid} activation 
to constrain the range of output values:
\begin{equation}
\begin{array}{l}
{y_{Rt}} = \tau \left( {W_{h{y_R}}}{h_t} + {b_{{y_R}}} \right),\\
{y_{et}} = \sigma \left( {{W_{h{y_e}}}{h_t} + {b_{{y_e}}}} \right),\\
{y_t} = \left( {{y_{Rt}},{y_{et}}} \right).
\end{array}
\label{eq:2output}
\end{equation}

We train the model by minimizing the squared error:
\begin{equation}
E = \sum\limits_t {{{\left\| {{y_t} - {{\hat y}_t}} \right\|}^2}},
\label{eq:loss}
\end{equation}
where ${\hat y}_t$ is the expected output, which we extract from training videos.
We use the CNTK deep learning toolkit to implement our neural network models.
Training hyperparameters are choosen as follows: minibatch size is 300, 
epoch size is 150,000 and learning rate is 0.0001. The network parameters
are learned by Adam optimizer~\cite{adam_iclr2015} in 300 epochs.

\section{EXPERIMENTS}
\label{sec:eval}

\begin{table*}[!ht]
	\centering
	\caption{RMSE of 3D landmarks in millimeter, categorized by types of emotions
		of test sequences, and by actors.}
	\label{tbl:3d_err}
	\begin{tabular}{|c|c||c|c|c|c|c|c|c|c||c|c|c|c|}
		\hline
		& \textbf{Mean} & Neutral & Calm & Happy & Sad & Angry & Fearful & Disgu. & Surpri. & Actor21 & Actor22 & Actor23 & Actor24 \\\hline
		LSTM~\cite{pham_cvprw2017} & 1.039 & 1.059 & 1.039 & 1.082 & 1.010 & 1.033 & 1.016 & 1.049 & 1.038 & 1.007 & 0.941 & 1.056 & 1.139 \\\hline
		CNN-static & 0.741 & 0.725 & 0.715 & 0.760 & 0.728 & 0.746 & 0.746 & 0.781 & 0.746 & 0.723 & 0.697 & 0.719 & 0.817 \\\hline
		CNN+LSTM & 1.042 & 1.077 & 1.029 & 1.092 & 1.029 & 1.013 & 1.031 & 1.035 & 1.054 & 1.026 & 0.946 & 1.021 & 1.162 \\\hline
		CNN+GRU & 1.022 & 1.034 & 0.995 & 1.081 & 0.999 & 1.012 & 1.008 & 1.023 & 1.045 & 0.998 & 0.952 & 0.985 & 1.139 \\\hline
	\end{tabular}
\end{table*}

\begin{table*}[!ht]
	\centering
	\caption{Mean squared error of expression blending weights, categorized by types of emotions
		of test sequences, and by actors.}
	\label{tbl:e_err}
	\begin{tabular}{|c|c||c|c|c|c|c|c|c|c||c|c|c|c|}
		\hline
		& \textbf{Mean} & Neutral & Calm & Happy & Sad & Angry & Fearful & Disgu. & Surpri. & Actor21 & Actor22 & Actor23 & Actor24 \\\hline
		LSTM~\cite{pham_cvprw2017} & 0.065 & 0.067 & 0.066 & 0.069 & 0.068 & 0.058 & 0.062 & 0.071 & 0.063 & 0.040 & 0.061 & 0.070 & 0.088 \\\hline
		CNN-static & 0.018 & 0.016 & 0.016 & 0.018 & 0.019 & 0.016 & 0.018 & 0.022 & 0.019 & 0.012 & 0.016 & 0.019 & 0.022 \\\hline
		CNN+LSTM & 0.074 & 0.072 & 0.074 & 0.083 & 0.074 & 0.063 & 0.073 & 0.074 & 0.079 & 0.052 & 0.061 & 0.075 & 0.106 \\\hline
		CNN+GRU & 0.067 & 0.065 & 0.065 & 0.075 & 0.069 & 0.059 & 0.065 & 0.073 & 0.070 & 0.041 & 0.061 & 0.066 & 0.100 \\\hline
	\end{tabular}
\end{table*}

\begin{table}[!ht]
	\centering
	\caption{Training errors after 300 epochs of four models.}
	\label{tbl:train_err}
	\begin{tabular}{|c|c|c|c|}
		\hline
		LSTM~\cite{pham_cvprw2017} & CNN-static & CNN+LSTM & CNN+GRU \\\hline
		0.59 & 0.94 & 0.54 & 0.52 \\\hline
	\end{tabular}
\end{table}

\subsection{Dataset}
\label{sec:data}
We use the Ryerson Audio-Visual Database of Emotional Speech and Song (RAVDESS)~\cite{ravdess} 
for training and evaluation. The database consists of 24 professional actors 
speaking and singing with various emotions: neutral, calm, happy, sad, angry, fearful, 
disgust and surprised. We use video sequences of the first 20 actors for training, 
with around 250,000 frames in total, which translate to about 2hr of audio,
and evaluate the model on the data of four remaining actors.

\subsection{Experimental Settings}
We train our deep neural network in two configurations: \textit{CNN+LSTM} and \textit{CNN+GRU},
in which the recurrent layer uses LSTM and GRU cells, respectively. As a baseline model, we drop
the recurrent layer in our proposed neural network, and denote it as \textit{CNN-static}. 
This model cannot handle smoothly temporal transition, 
it estimates facial parameters in a frame-by-frame basis. We also compared our proposed
models with the method described in~\cite{pham_cvprw2017}, which uses engineered features as input.

We compare these models on two metrics: RMSE of 3D landmark errors and mean squared error of facial
parameters, specifically, expression blending weights with respect to ground truth recovered by the
visual tracker~\cite{pham_icpr2016}. Landmark errors are calculated as distances
from the inner landmarks (showed in Figure~\ref{fig:face}) 
on the reconstructed 3D face shape, to those of the ground truth 3D shape.
We ignore error metrics for head rotation, 
since it is rather difficult to infer head pose correctly from the speech. We include head pose
estimation primarily to generate plausible rigid motions to augment the realism of the
talking avatar. However, these error metrics do not truly reflect the performance of our deep
generative model, based on our observations.

\subsection{Evaluation}
\label{sec:exper}

Table~\ref{tbl:3d_err} and~\ref{tbl:e_err} show the aforementioned error metrics
of four models, organized into different categories corresponding to
different emotions and testers. The proposed model with GRU cells slightly 
outperforms ~\cite{pham_cvprw2017} as well as the proposed model with LSTM cells, 
in terms of landmark errors. On the other hand, \textit{CNN+GRU} has similar blendshape 
coefficient errors to~\cite{pham_cvprw2017}, whereas \textit{CNN+LSTM} has highest errors.

\begin{figure}[!ht]
	\centering
	\includegraphics[width=8.5cm]{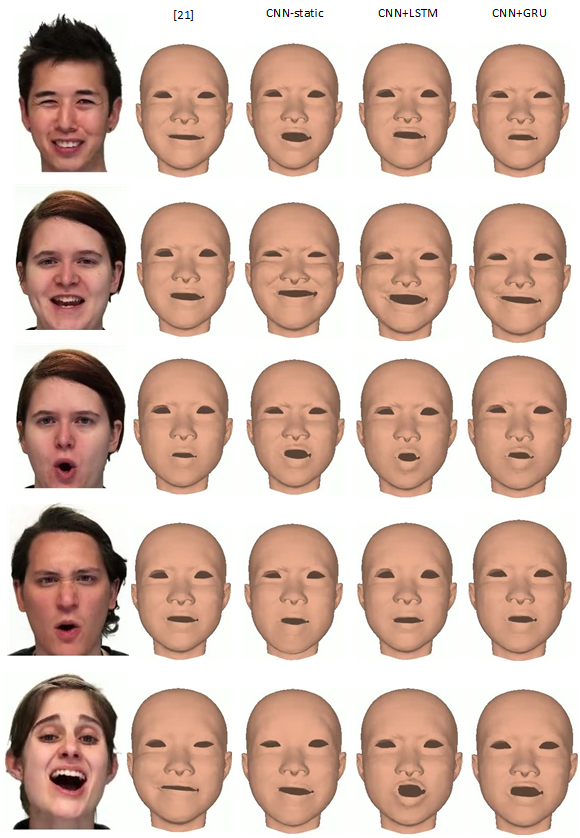}
	\caption{A few reconstruction samples. On the left are true face appearances.
	From the second to the last columns are reconstruction results 
	by Pham et al.~\cite{pham_cvprw2017}, \textit{CNN-static}, \textit{CNN+LSTM}
	and \textit{CNN+GRU}, respectively. We use a generic 3D face model
	animated with the parameters generated by each model. The reconstructions by
	\textit{CNN-static} depict emotions of the speakers reasonably well,
	however, it cannot generate smooth transition between frames.}
	\label{fig:exp} 
\end{figure}

Interestingly, on both metrics, the \textit{CNN-static}
outperforms all other models with recurrent layers. As shown in Table~\ref{tbl:3d_err},
the RMSE of \textit{CNN-static} is about 0.7mm consistently across different
categories and testers, and $30 \%$ lower than errors of other models. \textit{CNN-static}
also scores lower parameter estimation errors. These result suggests that the baseline
can actually generalizes well to test data in terms of facial action estimations, 
although visualization of 3D face reconstruction shows that the baseline model inherently generates
non-smooth sequence of facial actions, which is shown in the supplementary video. 
From these testing results, combined with
the training errors listed in Table~\ref{tbl:train_err}, we hypothesize that our proposed
models, \textit{CNN+LSTM} and \textit{CNN+GRU}, overfit the training data. And
\textit{CNN+GRU}, being simpler, can generalize somewhat better than \textit{CNN+LSTM}. 
The baseline model \textit{CNN-static}, being the simplest 
among the four i.e. it has the least number of parameters, 
generalizes well and achieves the best performance on the test set
in terms of error metrics, as well as emotional facial reconstruction from speech, 
demonstrated in Figure~\ref{fig:exp}.
These results once again prove the robustness, generalization and adaptability
of the CNN architecture and suggest that we have pursued the right
direction in using CNN to model facial action from raw waveforms, but there are
limitations and deficiencies in our current approach that need to be addressed.

\section{CONCLUSION AND FUTURE WORK}
\label{sec:conclude}
This paper introduces a deep learning framework for speech-driven 3D facial animation from raw
waveforms. Our proposed deep neural network learns a mapping from audio signal to the temporally varying
context of the speech, as well as emotional states of the speaker represented implicitly by blending weights
of a 3D face model.
Experiments demonstrate that our approach could estimate the form of lip movements with the emotional intensity 
of the speaker reasonably. However, there are certain limitations in our network architecture that 
prevent the model from reflecting the emotion in the speech perfectly. 
In the future, we will improve the generalization of our deep neural network,
and explore other generative models to increase the facial reconstruction quality.

\linespread{1.0}




\bibliographystyle{ieee}
\bibliography{mybib}

\begin{thebibliography}{10}\itemsep=-1pt

\bibitem{hamid_TASLP2014}
O.~Abdel-Hamid, A.-R. Mohamed, H.~Jiang, L.~Deng, G.~Penn, and D.~Yu.
\newblock Convolutional neural networks for speech recognition.
\newblock {\em IEEE Transaction on Audio, Speech, and Language Processing},
  22(10), October 2014.

\bibitem{hamid_icassp2012}
O.~Abdel-Hamid, A.-R. Mohamed, H.~Jiang, and G.~Penn.
\newblock Applying convolutional neural networks concepts to hybrid nn-hmm
  model for speech recognition.
\newblock In {\em IEEE International Conference on Acoustics, Speech and Signal
  Processing}, 2012.

\bibitem{blanz_03}
V.~Blanz, C.~Basso, T.~Poggio, and T.~Vetter.
\newblock Reanimating faces in images and video.
\newblock In {\em SIGGAPH}, pages 187--194, 1999.

\bibitem{blanz_99}
V.~Blanz and T.~Vetter.
\newblock A morphable model for the synthesis of 3d faces.
\newblock In {\em Eurographics}, pages 641--650, 2003.

\bibitem{bregler_siggraph07}
C.~Bregler, M.~Covell, and M.~Slaney.
\newblock Video rewrite: driving visual speech with audio.
\newblock In {\em SIGGRAPH}, pages 353--360, 2007.

\bibitem{cao_fwh2013gy}
C.~Cao, Y.~Weng, S.~Zhou, Y.~Tong, and K.~Zhou.
\newblock {FaceWarehouse: A 3D Facial Expression Database for Visual
  Computing}.
\newblock {\em IEEE Transactions on Visualization and Computer Graphics},
  20(3):413--425, March 2014.

\bibitem{yongcao_tog05}
Y.~Cao, W.~C. Tien, P.~Faloutsos, and F.~Pighin.
\newblock Expressive speech-driven facial animation.
\newblock {\em ACM Transactions on Graphics}, 24(4):1283--1302, 2005.

\bibitem{gru_chung_nipsw14}
J.~Chung, C.~Gulcehre, K.~Cho, and Y.~Bengio.
\newblock Empirical evaluation of gated recurrent neural networks on sequence
  modeling.
\newblock In {\em NIPS 2014 Deep Learning and Representation Learning
  Workshop}, 2014.

\bibitem{cosatto_03}
E.~Cosatto, J.~Ostermann, H.~P. Graf, and J.~Schroeter.
\newblock Lifelike talking faces for interactive services.
\newblock {\em Proc IEEE}, 91(9):1406--1429, 2003.

\bibitem{deng_icassp2013}
L.~Deng, O.~Abdel-Hamid, and D.~Yu.
\newblock A deep convolutional neural network using heterogeneous pooling for
  trading acoustic invariance with phonetic confusion.
\newblock In {\em IEEE International Conference on Acoustics, Speech and Signal
  Processing}, 2013.

\bibitem{ding_mta15}
C.~Ding, L.~Xie, and P.~Zhu.
\newblock Head motion synthesis from speech using deep neural network.
\newblock {\em Multimed Tools Appl}, 74:9871--9888, 2015.

\bibitem{ezzat_siggraph02}
T.~Ezzat, G.~Geiger, and T.~Poggio.
\newblock Trainable videorealistic speech animatio.
\newblock In {\em SIGGRAPH}, pages 388--397, 2002.

\bibitem{fan_mta16}
B.~Fan, L.~Xie, S.~Yang, L.~Wang, and F.~K. Soong.
\newblock A deep bidirectional lstm approach for video-realistic talking head.
\newblock {\em Multimed Tools Appl}, 75:5287--5309, 2016.

\bibitem{lstm_97}
S.~Hochreiter and J.~Scmidhuber.
\newblock Long short-term memory.
\newblock {\em Neural Comput}, 9(8):1735--1780, 1997.

\bibitem{hoshen_icassp2015}
Y.~Hoshen, R.~J. Weiss, and K.~W. Wilson.
\newblock Speech acoustic modeling from raw multichannel waveforms.
\newblock In {\em IEEE International Conference on Acoustics, Speech and Signal
  Processing}, 2015.

\bibitem{karras_sigg2017}
T.~Karras, T.~Aila, S.~Laine, A.~Herva, and J.~Lehtinen.
\newblock Audio-driven facial animation by joint end-to-end learning of pose
  and emotion.
\newblock In {\em SIGGRAPH}, 2017.

\bibitem{adam_iclr2015}
D.~P. Kingma and J.~Ba.
\newblock Adam: A method for stochastic optimization.
\newblock In {\em 3rd International Conference for Learning Representations},
  2015.

\bibitem{lecun_cnn}
Y.~LeCun and Y.~Bengio.
\newblock Convolutional networks for images, speech, and time-series.
\newblock pages 255--258, 1998.

\bibitem{ravdess}
S.~R. Livingstone, K.~Peck, and F.~A. Russo.
\newblock Ravdess: The ryerson audio-visual database of emotional speech and
  song.
\newblock In {\em 22nd Annual Meeting of the Canadian Society for Brain,
  Behaviour and Cognitive Science (CSBBCS)}, 2012.

\bibitem{palaz_is2013}
D.~Palaz, R.~Collobert, and M.~Magimai-Doss.
\newblock Estimating phoneme class conditional probabilities from raw speech
  signal using convolutional neural networks.
\newblock In {\em Interspeech}, 2013.

\bibitem{pham_cvprw2017}
H.~X. Pham, S.~Cheung, and V.~Pavlovic.
\newblock Speech-driven 3d facial animation with implicit emotional awareness:
  a deep learning approach.
\newblock In {\em The 1st DALCOM workshop, CVPR}, 2017.

\bibitem{pham_3dv2016}
H.~X. Pham and V.~Pavlovic.
\newblock Robust real-time 3d face tracking from rgbd videos under extreme
  pose, depth, and expression variations.
\newblock In {\em 3DV}, 2016.

\bibitem{pham_icpr2016}
H.~X. Pham, V.~Pavlovic, J.~Cai, and T.~jen Cham.
\newblock Robust real-time performance-driven 3d face tracking.
\newblock In {\em ICPR}, 2016.

\bibitem{qian_icassp14}
Y.~Qian, Y.~Fan, and F.~K. Soong.
\newblock On the training aspects of deep neural network (dnn) for parametric
  tts synthesis.
\newblock In {\em ICASSP}, pages 3829--3833, 2014.

\bibitem{sainath_NN2015}
T.~N. Sainath, B.~Kingsbury, G.~Saon, H.~Soltau, A.~rahman Mohamed, G.~Dahl,
  and B.~Ramabhadran.
\newblock Deep convolutional neural networks for large-scale speech tasks.
\newblock {\em Neural Network}, 64:39--48, 2015.

\bibitem{sainath_icassp2015}
T.~N. Sainath, O.~Vinyals, A.~Senior, and H.~Sak.
\newblock Convolutional, long short-term memory, fully connected deep neural
  networks.
\newblock In {\em IEEE International Conference on Acoustics, Speech and Signal
  Processing}, 2015.

\bibitem{sainath_is2015}
T.~N. Sainath, R.~J. Weiss, A.~Senior, K.~W. Wilson, and O.~Vinyals.
\newblock Learning the speech front-end with raw waveforms cldnns.
\newblock In {\em Interspeech}, 2015.

\bibitem{sako_icslp00}
S.~Sako, K.~Tokuda, T.~Masuko, T.~Kobayashi, and T.~Kitamura.
\newblock Hmm-based text-to-audio-visual speech synthesis.
\newblock In {\em ICSLP}, pages 25--28, 2000.

\bibitem{salvi_eurasip09}
G.~Salvi, J.~Beskow, S.~Moubayed, and B.~Granstrom.
\newblock Synface: speech-driven facial animation for virtual speech-reading
  support.
\newblock {\em URASIP journal on Audio, speech, and music processing}, 2009.

\bibitem{supasorn_sigg2017}
S.~Suwajanakorn, S.~M. Seitz, and I.~Kemelmacher-Schlizerman.
\newblock Synthesizing obama: learning lip sync from audio.
\newblock In {\em SIGGRAPH}, 2017.

\bibitem{taylor_sigg2017}
S.~Taylor, T.~Kim, Y.~Yue, M.~Mahler, J.~Krahe, A.~G. Rodriguez, J.~Hodgins,
  and I.~Matthews.
\newblock A deep learning approach for generalized speech animation.
\newblock In {\em SIGGRAPH}, 2017.

\bibitem{trigeorgis_is2016}
G.~Trigeorgis, F.~Ringeval, R.~Brueckner, E.~Marchi, M.~A. Nicolaou,
  B.~Schuller, and S.~Zafeiriou.
\newblock Adieu features? end-to-end speech emotion recognition using a deep
  convolutional recurrent network.
\newblock In {\em Interspeech}, 2016.

\bibitem{wang_sandbox07}
A.~Wang, M.~Emmi, and P.~Faloutsos.
\newblock Assembling an expressive facial animation system.
\newblock {\em ACM SIGGRAPH Video Game Symposium (Sandbox)}, pages 21--26,
  2007.

\bibitem{wang_interspk2010}
L.~Wang, X.~Qian, W.~Han, and F.~K. Soong.
\newblock Synthesizing photo-real talking head via trajectoryguided sample
  selection.
\newblock In {\em Interspeech}, pages 446--449, 2010.

\bibitem{wang_interspk11}
L.~Wang, X.~Qian, F.~K. Soong, and Q.~Huo.
\newblock Text driven 3d photo-realistic talking head.
\newblock In {\em Interspeech}, pages 3307--3310, 2011.

\bibitem{wu_interspk06}
Z.~Wu, S.~Zhang, L.~Cai, and H.~Meng.
\newblock Real-time synthesis of chinese visual speech and facial expressions
  using mpeg-4 fap features in a three-dimensional avatar.
\newblock In {\em Interspeech}, pages 1802--1805, 2006.

\bibitem{xie_multimed07}
L.~Xie and Z.~Liu.
\newblock Realistic mouth-synching for speech-driven talking face using
  articulatory modeling.
\newblock {\em IEEE Trans Multimed}, 9(23):500--510, 2007.

\bibitem{zen_icassp13}
H.~Zen, A.~Senior, and M.~Schuster.
\newblock Statistical parametric speech synthesis using deep neural networks.
\newblock In {\em ICASSP}, pages 7962--7966, 2013.

\bibitem{zhang_interspk13}
X.~Zhang, L.~Wang, G.~Li, F.~Seide, and F.~K. Soong.
\newblock A new language independent, photo realistic talking head driven by
  voice only.
\newblock In {\em Interspeech}, 2013.

\end{thebibliography}

\end{document}